# How Real Is AI Tutoring? Comparing Simulated and Human Dialogues in One-on-One Instruction

**Ruijia LI[a,b], Yuan-Hao JIANG[a,b], Jiatong WANG[a,b] & Bo JIANG[a,b*]**
[a]*Shanghai Institute of Artificial Intelligence for Education, East China Normal University, Shanghai, China*
[b]*Lab of Artificial Intelligence for Education, East China Normal University, Shanghai, China*
\*Corresponding author: bjiang@deit.ecnu.edu.cn

**Abstract:** Heuristic and scaffolded teacher-student dialogues are widely regarded as critical for fostering students' higher-order thinking and deep learning. However, large language models (LLMs) currently face challenges in generating pedagogically rich interactions. This study systematically investigates the structural and behavioral differences between AI-simulated and authentic human tutoring dialogues. We conducted a quantitative comparison using an Initiation-Response-Feedback (IRF) coding scheme and Epistemic Network Analysis (ENA). The results show that human dialogues are significantly superior to their AI counterparts in utterance length, as well as in questioning (I-Q) and general feedback (F-F) behaviors. More importantly, ENA results reveal a fundamental divergence in interactional patterns: **human dialogues are more cognitively guided and diverse**, centered around a "question-factual response-feedback" teaching loop that clearly reflects **pedagogical guidance and student-driven thinking**; in contrast, **simulated dialogues** exhibit a pattern of **structural simplification and behavioral convergence**, revolving around an "explanation-simplistic response" loop that is essentially a **simple information transfer** between the teacher and student. These findings illuminate key limitations in current AI-generated tutoring and provide empirical guidance for designing and evaluating more pedagogically effective generative educational dialogue systems.

## 1. Introduction

Heuristic and scaffolded teacher-student dialogues are widely recognized as a core pedagogical mechanism for promoting students' higher-order thinking and deep learning (Giuseffi, 2024). In particular, within one-on-one instructional settings, teachers support students' knowledge construction and cognitive development through questioning, responding, and providing feedback (Rivera et al., 2025). Prior research has demonstrated that structured instructional dialogues not only enhance students' conceptual understanding, but also offer observable linguistic evidence for learning and instructional analytics (Alisoy, 2025; Dai et al., 2025).

With the rapid advancement of Artificial Intelligence (AI) (Hong et al., 2025) and Large Language Models (LLMs) (OpenAI, 2023), AI-based dialogue systems have shown great potential for educational applications. From dialogic learning support to automated task generation, generative AI tools are offering unprecedented flexibility in supporting instruction. However, despite the impressive linguistic fluency of general-purpose LLMs, they still struggle to produce dialogues that are pedagogically guided and cognitively supportive (Joseph, 2023; Wu et al., 2025). Current models often fail to emulate the heuristic questioning of a teacher, the responsive behavior of a student, or the hierarchical and interactive structure of multi-turn tutoring. These limitations stem in part from a lack of multimodal understanding (Jiang et al., 2024; Mo, Shao, et al., 2025) and difficulties in faithfully reconstructing authentic human–AI interaction patterns (J. Chen et al., 2024; Mo, Huang, et al., 2025).

To enhance LLMs' instructional interaction capabilities, access to high-quality educational dialogue corpora is essential. However, real-world data collection faces challenges such as high cost (Qi et al., 2025), ethical restrictions (Duan, Shen, et al., 2024), structural inconsistencies and class imbalance (Duan, Gu, et al., 2024; Duan, Yang, et al., 2024), and significant noise in transcripts (Wang, Tu, et al., 2021; Wang, Zheng, et al., 2021), making such corpora less suitable as training or evaluation benchmarks (Zhang et al., 2023). Consequently, synthetic data generated by LLMs has emerged as a scalable and controllable alternative. Among these, the approach of prompting LLMs to simulate both the teacher and student roles in one-on-one dialogues is particularly promising. Nevertheless, the structural authenticity of these generated dialogues in comparison to human interactions remains an open empirical question (A. Chen et al., 2024).

This study focuses on one-on-one instructional scenarios, constructing two parallel corpora based on the same instructional prompts: one derived from authentic teacher-student dialogues, and the other generated through LLM simulations involving both roles. Using a coding framework grounded in the Initiation-Response-Feedback (IRF) structure (Y. Liu, 2008) and Epistemic Network Analysis (ENA) (Hila, 2025), we compare interactional patterns across the two corpora in terms of behavioral distributions and structural linkages. Findings reveal that authentic dialogues are characterized by greater cognitive nonlinearity, leapfrogging, and diversity, whereas simulated dialogues tend to converge toward simplified structures and repetitive patterns (Munday et al., 2023; Silseth & Furberg, 2024). To investigate the structural fidelity and cognitive pathways represented in AI-generated tutoring dialogues, this study addresses the following research questions:

- RQ1: What structural differences in instructional behaviors exist between AI-simulated and authentic one-on-one tutoring dialogues?
- RQ2: Do the structural linkages among behaviors in AI-simulated dialogues reflect distinct cognitive pathways and interactional patterns compared to those in authentic dialogues?

This study makes two primary contributions:

- It systematically compares AI-simulated and real-world one-on-one instructional dialogues under a unified task prompt, revealing critical shortcomings in LLMs' ability to produce interactive diversity and cognitive coherence.
- It proposes a dual-dimensional analysis framework combining IRF-based behavioral coding with ENA, offering a new paradigm for evaluating the structural quality of generative educational corpora.

## 2. Related Work

### 2.1 One-on-One Heuristic Dialogue Education

In individualized instructional contexts, heuristic dialogue has long been regarded as a fundamental mechanism supporting students' knowledge construction (Cavagnetto et al., 2010; Muhonen et al., 2017). Vygotsky's theory of the Zone of Proximal Development emphasizes the role of guided interaction in helping learners reach their developmental potential (Vygotsky & Cole, 1978). Mercer introduced the notion of "exploratory talk," highlighting the importance of collaborative and reflective discourse in promoting conceptual internalization and cognitive expansion (Mercer, 2002). Alexander further proposed the framework of "dialogic teaching," which values authentic, cumulative, and reflective classroom interaction (Alexander, 2008). More recently, Socratic questioning has gained traction as a heuristic instructional tool aimed at stimulating learners' reflection through strategic inquiry (Qi et al., 2024). Within this theoretical backdrop, the IRF (Initiation–Response–Feedback) structure has been widely adopted as a classical model for analyzing instructional dialogue. It is particularly effective in visualizing and interpreting the cognitive support pathways and pedagogical pacing inherent in one-on-one educational settings (Waring, 2008, 2009). In this study, we adopt IRF-based behavioral coding to analyze and compare authentic and synthetic dialogues, offering empirical insights into the potential of LLM-driven heuristic dialogue education.

### 2.2 AI and the Simulation of Dialogue Education

With the rapid advancement of Generative AI (GenAI) technologies (Jiang, Wei, et al., 2025;

Wei et al., 2024), LLMs, AI agents, and agentic workflows are increasingly applied in instructional resource generation (Li et al., 2024), teacher behavior simulation (Jiang, Chen, et al., 2025), and corpus construction for educational applications (Zhuang, Mao, et al., 2024). These approaches offer a potential solution to the shortage of high-quality educational data, yet also raise critical questions regarding the authenticity and pedagogical value of AI-generated content. Prior studies have explored LLMs' ability to generate teacher questions (Shi et al., 2023), instructional feedback (Zhuang, Wu, et al., 2024), and simulations of knowledge component transitions during learning (Jiang, Tang, et al., 2025). Nonetheless, many of these systems suffer from rigid structure, homogenized dialogue acts, and shallow cognitive engagement (Cuskley et al., 2024; Niu et al., 2024). These issues bring us to a key question: Do simulated dialogues reflect the structural and interactional features of real instructional exchanges? This study seeks to empirically address this question and provide methodological guidance for the development and evaluation of generative educational corpora.

### 3. Methods

*3.1 Participants and experiment procedure*

The study involved fifth-grade students from an elementary school in Xundian County, Yunnan Province, China. Undergraduate and graduate volunteers from a university in eastern China conducted an eight-week mathematics tutoring program in a one-on-one online format. Prior to the tutoring sessions, the volunteers received specialized training in the Socratic questioning technique, aiming to foster students' ability to construct their own understanding frameworks and develop independent problem-solving skills, rather than simply providing direct answers. The tutoring content primarily covered two sections of the elementary school mathematics curriculum: "Numbers and Algebra" and "Geometry and Graphics." All teacher–student interactions during the experiment were recorded via the Tencent Meeting platform, which automatically generated transcripts of the recorded sessions. During data processing, any transcription errors or disfluent expressions were refined sentence by sentence using GPT-based text polishing. After data cleaning and screening, 49 dialogues corresponding to mathematics problems were retained as the final research corpus.

*3.2 Simulation Data Generation*

To explore the differences between real conversations and AI-simulated conversations in terms of teaching methods and dialogue structures, while minimizing the influence of problem statements and solution outlines on the dialogue itself, this study first extracts the tutoring question from the original dialogue and distills the core tutoring approach. This serves as both the basis and the constraints for the simulation, preventing the dialogue from diverging excessively (Figure 1). Using the question and core tutoring approach as inputs, the study adopts the tripartite simulation framework proposed by Liu et al., (2024) in SocraticLM, which has been shown to generate data suitable for fine-tuning large language models to enhance their Socratic questioning capabilities, making it a valuable methodological reference.

In this framework, three AI agents collaboratively simulate a Socratic teaching scenario: (1) the Teacher agent generates heuristic questions based on predefined instructional objectives; (2) the student agent responds according to individualized knowledge gaps; and (3) the Dean agent monitors dialogue quality, determines turn-taking, and ensures instructional coherence. The entire simulation is implemented on the GPT-4o model using a predefined prompt template.

*3.3 Coding process and framework*

This study is based on the IRF (Initiation-Response-Feedback) discourse framework proposed by Sinclair & Coulthard (2013), a widely recognized model for analyzing dialogue in educational settings. To systematically examine one-on-one dialogue quality, we adopted a refined coding scheme for each dimension of the IRF model (Table 1). The sub-categories for both the Initiation (I) and Feedback (F) dimensions are adapted from the six teacher scaffolding types identified by Van de Pol et al., (2010). Based on our research approach, we conceptualize three of these behaviors—Questioning (I-Q), Hints (I-H), and Modeling (I-M)—as typically occurring when the teacher initiates a dialogue, thus placing them in the Initiation dimension. The Response (R) dimension, in turn, builds upon Yang et al. (2023)'s hierarchical classification, with an expansion to suit this study's context, which includes a Simplistic Response (R-SR), a Factual Response (R-FR), an Open-ended Response (R-IO), and the newly added Refusal to Respond (R-RR). Finally, the Feedback (F) dimension includes the remaining three scaffolding types—Feeding Back (F-F), Instructing (F-I), and Explaining (F-E)—which are used to code the teacher's follow-up actions to a student's response. This comprehensive coding framework provides a detailed analytical tool to explore the complexity and quality of one-on-one dialogue within the present study.

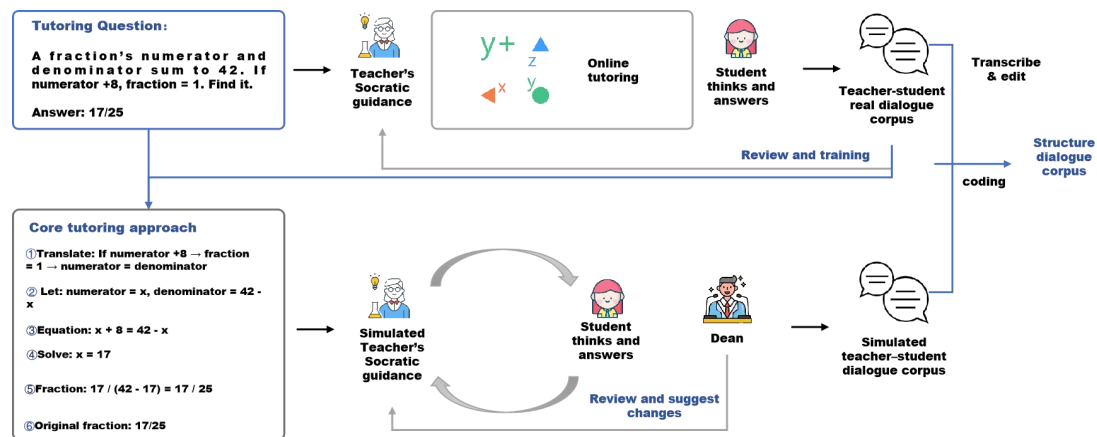

*Figure 1*. Workflow of dialogue simulation

### 3.4 Data analysis

This study first applied the coding scheme from Section 3.3 to both the human-to-human dialogues collected in Section 3.1 and the AI-simulated dialogues from Section 3.2. The human-to-human dialogues were coded independently by two human researchers, achieving a high inter-rater reliability with a Cohen's Kappa of 0.824. Subsequently, the expert-coded human dialogue data were used to fine-tune a BERT model for the purpose of automatic dialogue coding. This fine-tuned BERT model was then used to automatically code the AI-simulated dialogues, with manual verification to ensure accuracy.

To uncover the underlying differences between the AI-simulated and human-to-human dialogues, a multi-faceted analytical approach was employed. Initially, we used descriptive statistics and paired-samples T-tests to analyze the structural differences between the two dialogue sets at the IRF level, followed by a deeper analysis of the differences in cognitive depth within the sub-dimensions. Beyond these comparative statistics, the study further utilized Epistemic Network Analysis (ENA) to provide a more profound insight into the structural disparities. ENA is a novel method for quantifying and visualizing the relationships between elements within a discourse. In this study, we used the ENA Web Toolkit to perform network analysis on both dialogue sets, aiming to capture the potential differences in interaction patterns between AI-simulated and human dialogues at a macro-structural level.

## 4. Results

### 4.1 RQ1: Dialogue Structural Differences

To explore the structural differences between AI-simulated and authentic human dialogues, we conducted a quantitative analysis across two key dimensions: utterance length and the composition of instructional behaviors. Our analysis, as illustrated in Figure 2 (left), reveals significant structural disparities in utterance length between AI and human dialogues. Highly significant differences were found in both teacher and student roles ($p<.001$). Specifically, authentic human dialogue exhibits a dynamic and asymmetrical pattern of utterance length between different roles, whereas AI dialogue demonstrates a more uniform and standardized characteristic. Human teachers tend to produce longer utterances, while human students respond with very short turns, a pattern less pronounced in the AI-simulated interactions.

**Table 1.** *IRF-based conversation coding framework*

| Type | Subtype | Definition | Example |
|---|---|---|---|
| **I-Initiation** | I-Q Questioning | Involves asking students questions that require an active linguistic and cognitive answer. | What is 5+3? How did you get the answer? |
| | I-H Hints | The teacher provides clues or suggestions to help the student progress, without giving the full solution. | We need to find the grandfather's age. We know the age difference between the father and Tom, and between the grandfather and the father. How can we use this information? |
| | I-M Modeling | The process of offering a behavior for imitation, such as demonstrating a skill. | I will show you how to solve this vertical equation. |
| **R-Response** | R-RR Refuse to Response | Refusing to answer or remaining silent. | "I don't know," or silence. |
| | R-SR Simplistic Response | A simple answer that lacks depth. | "Yeah," "mm," or "okay." |
| | R-FR Factual Response | An answer that is factual, based on memory, or explanatory. | The sum of a triangle's interior angles is 180 degrees. |
| | R-IO Interpretive/Open-ended | A thorough answer that includes interpretation or explanation. | I think we should first find the location of 1, and then calculate the coordinates for 1/3. |
| **F-Feedback** | F-F Feeding Back | Providing information to the student about their own performance. | "Good job," or "That's smart." |
| | F-I Instructing | The teacher tells the student what to do or explains how/why something must be done. | Tom is 5, and the age difference is 30, so the father is 35. The grandfather is 30 years older, so his age is 35+30. |
| | F-E Explaining | The teacher provides more detailed information or clarification. | The sum of a triangle's angles is 180°. Since the other two angles are 30° and 60°, the third angle must be 90°. |

The composition of instructional behaviors, shown in Figure 2 (right), further highlights these differences. The proportion of Initiation (I) codes was significantly higher in human dialogues compared to AI ($T=-2.39$, $p<.05$). This suggests that human tutors are more inclined to use directive and guiding language, as indicated by the higher distribution of 'I' codes for the 'human' condition. However, no statistically significant differences were observed in the proportions of Response (R) and Feedback (F) codes. This indicates that the AI system is effective at replicating these fundamental interactive and evaluative behaviors, as the distributions for 'R' and 'F' codes appear largely similar between human and AI.

These findings suggest that while AI dialogue can successfully mimic certain human interaction patterns, as seen in the 'R' and 'F' code proportions, it still fundamentally differs in discourse granularity and the proactive nature of instructional initiation. Authentic human tutors exhibit a unique capacity for proactive guidance through longer and more frequent initiations, while AI dialogues tend to adopt a more reactive and standardized communication style. Nevertheless, the AI system's performance in basic interactional functions provides a clear direction for future improvements, particularly in enhancing the richness and diversity of its directive behaviors to better align with human tutoring strategies.

To gain a more granular understanding of the source of these differences, we conducted a paired samples t-test on the subtypes of instructional behaviors, with the results shown in Table 2. This detailed analysis reveals fundamental distinctions in the specific behavioral patterns between AI and human dialogues, moving beyond mere proportional differences.

First, regarding dialogue initiation, human tutors are more inclined to guide students through a higher frequency of Questioning (I-Q) ($p<.01$). This behavioral pattern reflects a common cognitive scaffolding strategy in human teaching, where continuous inquiry, rather than direct information delivery, is used to facilitate students' active knowledge construction. In contrast, although AI showed no significant difference from humans in behaviors like Hints (I-H) and Modeling (I-M), its deficiency in questioning may render its deep cognitive guidance more passive.

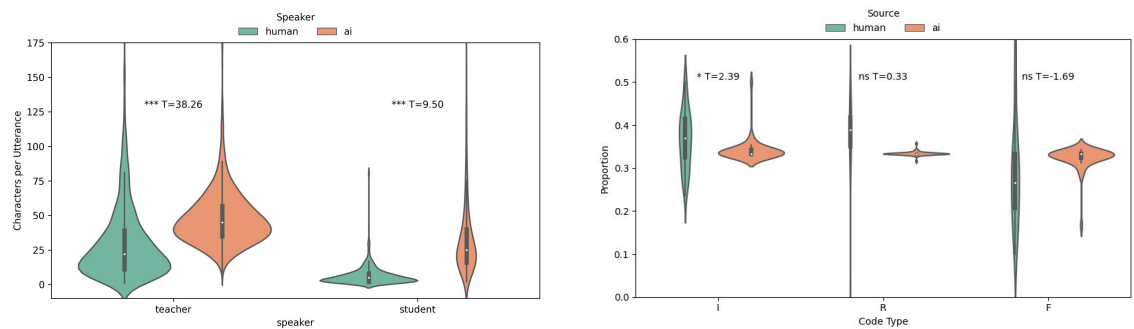

*Figure 2.* Comparison of dialogue structural characteristics between human and AI-simulated tutoring sessions

Second, these differences are mirrored in student response behaviors. Human students more frequently provide Factual Responses (R-FR) ($p<.001$), which might stem from the human tutor's persistent questioning strategy that prompts students to recall and articulate factual knowledge. In stark contrast, AI students are more prone to giving Simplistic Responses (R-SR) ($p<.001$) and Refused Responses (R-RR) ($p<.05$). This pattern may suggest that AI students' design favors providing direct, concise answers when possible, or adopting a "refusal" strategy when a full response is not feasible, unlike human students who attempt to organize facts to construct a response.

Finally, in feedback behaviors, AI demonstrates a unique set of strengths and limitations. AI tutors show a significant advantage in Explaining (F-E) ($p<.001$), with a much higher proportion than human tutors. This is likely an inherent capability of AI as an information generation system, allowing it to efficiently and systematically produce detailed explanations. In contrast, human tutors more often provide general Feedback (F-F) ($p<.01$), which may include immediate, unstructured evaluations of a student's effort or overall performance—a nuance AI currently struggles to fully replicate. These fine-grained analyses suggest that the differences between AI and human dialogues are not coincidental but are driven by fundamental distinctions in their underlying mechanisms and behavioral strategies. AI dialogue leans more towards information transfer and task completion, while human dialogue is more focused on cognitive guidance and social-emotional interaction.

### 4.2 RQ2：Comparison of Cognitive and Interactional Patterns

To investigate the deeper structural differences in interactive patterns between AI-simulated and authentic human dialogues, we employed Epistemic Network Analysis (ENA). This analysis aims to reveal how different instructional behaviors co-occur and the underlying cognitive and interactional patterns they collectively constitute. Our ENA results (as shown in Figure 3) reveal a significant structural divergence between the two groups. The centroids of the human and AI dialogue networks are significantly separated along the X-axis of the ENA projection space ($t(84.35)=9.33$, $p<0.001$, $d=1.97$), while no significant difference was found along the Y-axis. This suggests that the fundamental distinction between AI and human dialogues primarily resides within a single, core dimension.

A closer examination of each group's network structure reveals that the X-axis represents the difference between question-centered, guided instruction and explanation-centered, information-transfer instruction. The human dialogue network (red network) is structured around a strong connection between I-Q (Questioning) and R-FR (Factual Response) (as shown in Figure 4, left panel). This portrays a typical "question-factual response-feedback" teaching cycle, reflecting a pedagogical style where human tutors use Socratic questioning to drive knowledge retrieval and construction. In contrast, the AI dialogue network (blue network) exhibits a distinctly different pattern. The strongest connections are found between F-E (Explaining) and R-SR (Simplistic Response), depicting an "explanation-simplistic response" loop (as shown in Figure 4, right panel). This suggests that the AI tutor's strategy leverages its strength as an information repository, providing detailed explanations to deliver information, while the student responds with brief, confirmatory utterances.

Table 2. *Paired samples t-test results for AI vs. human dialogue behaviors by subtype*

| code | ai_mean | ai_std | human_mean | human_std | t_stat | p_value |
|---|---|---|---|---|---|---|
| I-H | 0.038 | 0.048 | 0.025 | 0.034 | 1.407 | 0.163 |
| I-Q | 0.302 | 0.043 | 0.339 | 0.068 | -3.121 | 0.002** |
| I-M | 0.002 | 0.010 | 0.003 | 0.021 | -0.534 | 0.595 |
| R-SR | 0.223 | 0.081 | 0.147 | 0.083 | 4.356 | <0.001*** |
| R-FR | 0.083 | 0.082 | 0.178 | 0.093 | -5.163 | <0.001*** |
| R-IO | 0.005 | 0.017 | 0.008 | 0.020 | -0.872 | 0.386 |
| R-RR | 0.024 | 0.039 | 0.008 | 0.011 | 2.591 | 0.012* |
| F-I | 0.169 | 0.078 | 0.138 | 0.099 | 1.604 | 0.112 |
| F-F | 0.075 | 0.093 | 0.136 | 0.087 | -3.161 | 0.002** |
| F-E | 0.080 | 0.074 | 0.016 | 0.030 | 5.382 | <0.001*** |

In summary, these findings collectively address our research question: the interactional patterns of AI-simulated dialogues fundamentally differ from those of authentic human dialogues. Human dialogue is more guided and interactive, promoting cognitive construction, whereas AI dialogue is more information-transfer-oriented and task-driven, focused on efficient information exchange. The ENA results not only confirm this difference but also precisely reveal the distinct cognitive and pedagogical pathways underlying these two modes of dialogue.

## 5. Discussion & Conclusion

Our analysis reveals a fundamental difference in the interaction patterns of human and AI tutors. Human tutors predominantly utilized a "question-response-feedback" loop, a method consistent with Cognitive Scaffolding theory that actively guides students in knowledge construction and promotes critical thinking. In contrast, the AI agent adopted an "explanation-simplistic response" pattern, functioning more as an efficient information-transfer mechanism than a pedagogical partner. This suggests that while current AI can achieve conversational fluency, it struggles to replicate the deeper, heuristic interactions essential for robust learning.

The main contributions of this work are threefold. First, we provide quantitative evidence that AI and human tutoring dialogues differ significantly in their behavioral composition and interactional structure. Second, we clarify the distinction between the "cognitive guidance" pathway characteristic of human tutors and the "information transfer" pathway currently favored by AI. Third, this study offers an empirical framework for designing and evaluating future AIED systems, advocating for a shift in focus from mere fluency to pedagogical authenticity.

These conclusions, however, should be considered in light of key methodological limitations. Our human tutor sample consisted of university students rather than seasoned educators; the simulated student agent did not capture the cognitive diversity of real learners; and the AI's behavior is an entangled product of the underlying LLM and our specific agent design. These constraints directly inform directions for future research. Subsequent work should extend this analysis to expert educators, incorporate more varied learner profiles into simulations, and explore novel training methods that enable AI to better emulate the complex scaffolding and heuristic behaviors found in authentic human dialogue.

## Acknowledgements

This work was partially supported by the National Natural Science Foundation of China, under the Grant 62477012, and the Natural Science Foundation of Shanghai, under the Grant 23ZR1418500, and the Special Foundation for Interdisciplinary Talent Training in "AI Empowered Psychology / Education" of the School of Computer Science and Technology, East China Normal University, under the Grant 2024JCRC-03. This work was also jointly supported in part by the Fundamental Research Funds for the Central Universities, and the ECNU Academic Innovation Promotion Program for Excellent Doctoral Students (both under Grant YBNLTS2025-008).

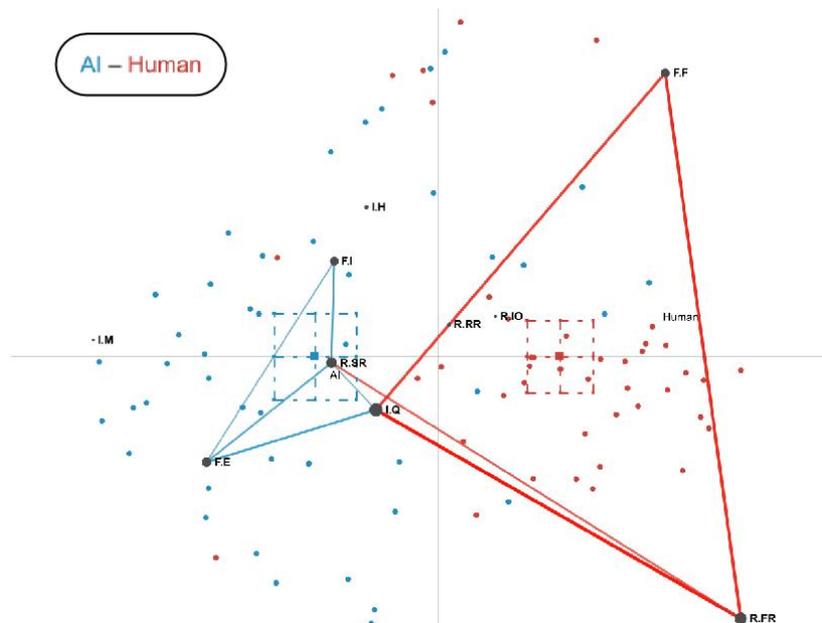

*Figure 3.* Comparison of dialogue network structures in AI-simulated and human tutoring sessions

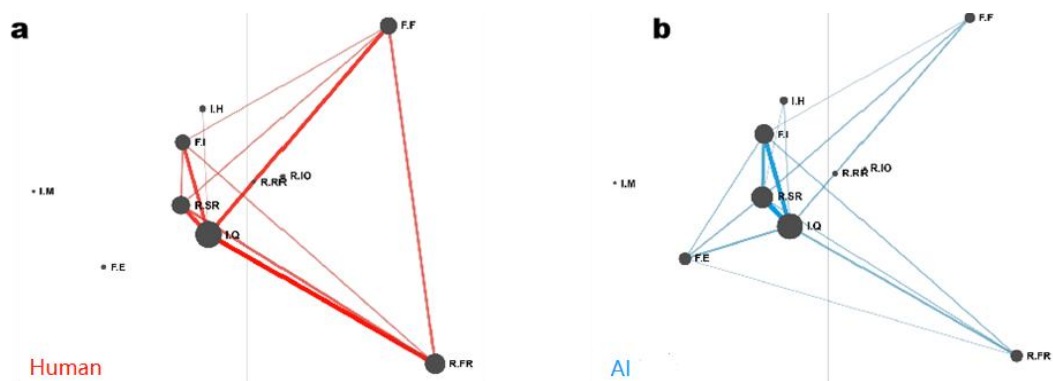

*Figure 4.* Distinct interactional patterns of human and AI dialogues

## References

Alexander, R. J. (2008). *Towards dialogic teaching: Rethinking classroom talk*.
Alisoy, H. (2025). From Echo Chambers to Critical Dialogue: A Comparative Case Study of Social Media-Based Pedagogy For Addressing Scientific Misinformation. *Global Spectrum of Research and Humanities*, *2*(4), 35–47. https://doi.org/10.69760/gsrh.0250203002
Cavagnetto, A., Hand, B. M., & Norton-Meier, L. (2010). The nature of elementary student science discourse in the context of the science writing heuristic approach. *International Journal of Science Education*, *32*(4), 427–449.
Chen, A., Wei, Y., Le, H., & Zhang, Y. (2024). *Learning-by-teaching with ChatGPT: The effect of teachable ChatGPT agent on programming education* (No. arXiv:2412.15226). arXiv.
Chen, J., Dai, L., Yuan-Hao, J., Zhou, Y., & Kong, X. (2024). Human-Computer Collaboration: A New Paradigm of Conversational Learning Based on ChatGPT. In Y. Wei, C. Qi, Y.-H. Jiang, & L. Dai (Eds.), *Enhancing Educational Practices: Strategies for Assessing and Improving Learning Outcomes* (pp. 209–225). Nova Science Publishers. https://doi.org/10.52305/RUIG5131

Cuskley, C., Woods, R., & Flaherty, M. (2024). The limitations of large language models for understanding human language and cognition. *Open Mind*, *8*, 1058–1083.

Dai, L., Jin, W., Zhu, B., Liao, R., Xu, G., Jiang, H., & Guan, J. (2025). Exploring the role of social media in mathematics learning: Effects on self-efficacy, interest, and self-regulation. *BMC Psychology*, *13*(1), 829.

Duan, J., Gu, Y., Yu, H., Yang, X., & Gao, S. (2024). ECC + +: An algorithm family based on ensemble of classifier chains for classifying imbalanced multi-label data. *Expert Systems with Applications*, *236*, 121366. https://doi.org/10.1016/j.eswa.2023.121366

Duan, J., Shen, H.-J., Duan, B.-M., Wang, Q., & Jiang, Y.-H. (2024). EFDR-CI: An Ensemble Learning Framework for Assessing Dropout Risk in the Context of Class Imbalance. In Y. Wei, C. Qi, Y.-H. Jiang, & L. Dai (Eds.), *Enhancing Educational Practices: Strategies for Assessing and Improving Learning Outcomes* (pp. 189–206). Nova Science Publishers.

Duan, J., Yang, X., Gao, S., & Yu, H. (2024). A partition-based problem transformation algorithm for classifying imbalanced multi-label data. *Engineering Applications of Artificial Intelligence*, *128*, 107506.

Giuseffi, F. (2024). The Investigation of a Nelsonian Approach to Socratic Dialogue with Student-Teachers at a Midwestern Private University. *InSight: A Journal of Scholarly Teaching*, *20*.

Hila, A. (2025). The epistemological consequences of large language models: Rethinking collective intelligence and institutional knowledge. *AI & SOCIETY*.

Hong, H., Dai, L., & Zheng, X. (2025). Advances in Wearable Sensors for Learning Analytics: Trends, Challenges, and Prospects. *Sensors*, *25*(9), Article 9. https://doi.org/10.3390/s25092714

Jiang, Y.-H., Chen, Z.-W., Zhao, C., Tang, K., Duan, J., & Zhou, Y. (2025). Explainable Learning Outcomes Prediction: Information Fusion Based on Grades Time-Series and Student Behaviors. *Proceedings of the Extended Abstracts of the CHI Conference on Human Factors in Computing Systems*, 1–11. https://doi.org/10.1145/3706599.3721212

Jiang, Y.-H., Li, R., Wei, Y., Jia, R., Shao, X., Hu, H., & Jiang, B. (2024). Synchronizing Verbal Responses and Board Writing for Multimodal Math Instruction with LLMs. *NeurIPS'24: Conference and Workshop on Neural Information Processing Systems, the 4th Workshop on Mathematical Reasoning and AI*, 46–59. https://openreview.net/forum?id=esbIrV8N12

Jiang, Y.-H., Tang, K., Chen, Z.-W., Wei, Y., Liu, T.-Y., & Wu, J. (2025). MAS-KCL: Knowledge component graph structure learning with large language model-based agentic workflow. *The Visual Computer*, *41*(2025), 6453–6464. https://doi.org/10.1007/s00371-025-03946-1

Jiang, Y.-H., Wei, Y., Shao, X., Jia, R., Zhou, Y., & Chen, Z.-W. (2025). *Generative AI in Personalized Learning: Development Trajectory, Educational Applications, and Future Education*. 710–719. https://www.learntechlib.org/primary/p/225586/

Joseph, S. (2023). Large Language Model-based Tools in Language Teaching to Develop Critical Thinking and Sustainable Cognitive Structures. *Rupkatha Journal on Interdisciplinary Studies in Humanities*, *15*(4).

Li, R., Jiang, Y.-H., Wang, Y., Hu, H., & Jiang, B. (2024). A Large Language Model-Enabled Solution for the Automatic Generation of Situated Multiple-Choice Math Questions. *Conference Proceedings of the 28th Global Chinese Conference on Computers in Education (GCCCE 2024)*, 130–136. http://gccce2024.swu.edu.cn/GCCCE2024_gongzuofanglunwenji2024-06-23A.pdf#page=148

Liu, J., Huang, Z., Xiao, T., Sha, J., Wu, J., Liu, Q., Wang, S., & Chen, E. (2024). SocraticLM: Exploring socratic personalized teaching with large language models. *Advances in Neural Information Processing Systems*, *37*, 85693–85721.

Liu, Y. (2008). Teacher–student talk in Singapore Chinese language classrooms: A case study of initiation/response/follow-up (IRF). *Asia Pacific Journal of Education*, *28*(1), 87–102.

Mercer, N. (2002). *Words and minds: How we use language to think together*. Routledge.

Mo, Y., Huang, G., li, liangcheng, Deng, D., Yu, Z., Xu, Y., Ye, K., Zhou, S., & Bu, J. (2025). TableNarrator: Making Image Tables Accessible to Blind and Low Vision People. *Proceedings of the 2025 CHI Conference on Human Factors in Computing Systems*, 1–17.

Mo, Y., Shao, Z., Ye, K., Mao, X., Zhang, B., Xing, H., Ye, P., Huang, G., Chen, K., Huan, Z., Yan, Z., & Zhou, S. (2025). *Doc-CoB: Enhancing Multi-Modal Document Understanding with Visual Chain-of-Boxes Reasoning* (No. arXiv:2505.18603). arXiv.

Muhonen, H., Rasku-Puttonen, H., Pakarinen, E., Poikkeus, A.-M., & Lerkkanen, M.-K. (2017). Knowledge-building patterns in educational dialogue. *International Journal of Educational Research*, *81*, 25–37.

Munday, I., Heinz, M., & Gallagher, B. (2023). Reflections on Distance in Remote Placement Supervision: Bodies, Power, and Negative Education. *Education Sciences*, *14*(1), 5.

Niu, Q., Liu, J., Bi, Z., Feng, P., Peng, B., Chen, K., Li, M., Yan, L. K., Zhang, Y., Yin, C. H., & others. (2024). Large language models and cognitive science: A comprehensive review of similarities, differences, and challenges. *arXiv Preprint arXiv:2409.02387*.

OpenAI, R. (2023). GPT-4 technical report. Arxiv 2303.08774. *View in Article*, *2*, 3.

Qi, C., Jia, L., Wei, Y., Jiang, Y.-H., & Gu, X. (2024). IntelliChain: An Integrated Framework for Enhanced Socratic Method Dialogue with LLMs and Knowledge Graphs. *Conference Proceedings of the 28th Global Chinese Conference on Computers in Education (GCCCE 2024)*, 116–121. https://doi.org/10.48550/arXiv.2502.00010

Qi, C., Zheng, L., Wei, Y., Xu, H., Chen, P., & Gu, X. (2025). EduDCM: A Novel Framework for Automatic Educational Dialogue Classification Dataset Construction via Distant Supervision and Large Language Models. *Applied Sciences*, *15*(1), Article 1.

Rivera, D. A., Frenay, M., Paquot, M., De Montpellier, P., & Swaen, V. (2025). Beyond the process: A novel analytical model to examine knowledge construction in MOOC forums. *Computers & Education*, *235*, 105342. https://doi.org/10.1016/j.compedu.2025.105342

Shi, J., Zhao, J., Wang, Y., Wu, X., Li, J., & He, L. (2023). CGMI: Configurable general multi-agent interaction framework. *arXiv Preprint arXiv:2308.12503*.

Silseth, K., & Furberg, A. (2024). Bridging group work and whole-class activities through responsive teaching in science education. *European Journal of Psychology of Education*, *39*(3), 2155–2176.

Sinclair, J., & Coulthard, M. (2013). Towards an analysis of discourse. In *Advances in spoken discourse analysis* (pp. 1–34). Routledge.

Van de Pol, J., Volman, M., & Beishuizen, J. (2010). Scaffolding in teacher–student interaction: A decade of research. *Educational Psychology Review*, *22*(3), 271–296.

Vygotsky, L. S., & Cole, M. (1978). *Mind in society: Development of higher psychological processes*. Harvard university press.

Wang, W., Tu, Y., Song, L., Zheng, J., & Wang, T. (2021). An Adaptive Design for Item Parameter Online Estimation and Q-Matrix Online Calibration in CD-CAT. *Frontiers in Psychology*, *12*.

Wang, W., Zheng, J., Song, L., Tu, Y., & Gao, P. (2021). Test Assembly for Cognitive Diagnosis Using Mixed-Integer Linear Programming. *Frontiers in Psychology*, *12*.

Waring, H. Z. (2008). Using explicit positive assessment in the language classroom: IRF, feedback, and learning opportunities. *The Modern Language Journal*, *92*(4), 577–594.

Waring, H. Z. (2009). Moving out of IRF (Initiation-Response-Feedback): A single case analysis. *Language Learning*, *59*(4), 796–824.

Wei, Y., Qi, C., Jiang, Y.-H., & Dai, L. (Eds.). (2024). *Enhancing Educational Practices: Strategies for Assessing and Improving Learning Outcomes*. Nova Science Publishers. https://doi.org/10.52305/RUIG5131

Wu, S., Oltramari, A., Francis, J., Giles, C. L., & Ritter, F. E. (2025). LLM-ACTR: From Cognitive Models to LLMs in Manufacturing Solutions. *Proceedings of the AAAI Symposium Series*, *5*(1), 340–349.

Yang, X., Wang, Q., & Jiang, J. (2023). Analysis of classroom teacher-student dialogue based on artificial intelligence:automatic classification and sub-level construction of IRE. *E-Education Research*, *44*(10), 79–86. https://doi.org/10.13811/j.cnki.eer.2023.10.011

Zhang, J., Shen, Z., Yang, S., Meng, L., Xiao, C., Jia, W., Li, Y., Sun, Q., Zhang, W., & Lin, X. (2023). High-Ratio Compression for Machine-Generated Data. *Proc. ACM Manag. Data*, *1*(4), 245:1-245:27. https://doi.org/10.1145/3626732

Zhuang, X., Mao, X., Jiang, Y.-H., Wu, H., Zhao, S., Cai, L., Liu, S., Chen, Y., Song, Y., Jia, C., Zhou, Y., & Lan, M. (2024). Bread: A Hybrid Approach for Instruction Data Mining through Balanced Retrieval and Dynamic Data Sampling. *NLPCC 2024*, 229–240.

Zhuang, X., Wu, H., Shen, X., Yu, P., Yi, G., Chen, X., Hu, T., Chen, Y., Ren, Y., Zhang, Y., Song, Y., Liu, B., & Lan, M. (2024). TOREE: Evaluating Topic Relevance of Student Essays for Chinese Primary and Middle School Education. In *ACL Findings 2024* (pp. 5749–5765). ACL.